\documentclass{article}
\usepackage[margin=1in]{geometry}
\usepackage[dvipsnames]{xcolor}
\usepackage{booktabs}
\usepackage{graphicx}
\usepackage{subcaption}
\usepackage{amsmath,amssymb}
\usepackage{cleveref}
\usepackage{algorithm} 
\usepackage{algpseudocode} 
\usepackage{calc}
\usepackage{tabto}
\usepackage{comment}
\usepackage[utf8]{inputenc}
\usepackage[english]{babel}
\usepackage{subcaption}
\usepackage{lipsum, color}
\usepackage{blindtext}

\usepackage{amsmath,amsthm}
\usepackage{mathtools, cuted}
\usepackage{wrapfig}

\title{\LARGE \bf
Online Tool Selection with Learned Grasp Prediction Models
}
\author{\small Khashayar Rohanimanesh$^{1}$, Jake Metzger$^{2}$, William Richards$^{1}$, and Aviv Tamar$^{3}$% <-this % stops a space
\thanks{$^{1}$Osaro Inc., 500 3rd street, San Francisco, CA 94103, USA
        {\tt\small \{khash,will\}@osaro.com}}%
\thanks{$^{2}$Accenture Inc., 415 Mission street, San Francisco, CA 94105, USA
        {\tt\small jacob.c.metzger@accenture.com}}%
\thanks{$^{3}$Technion Institute of Technology, 458 Fischbach building, Technion
Haifa 32000
Israel
        {\tt\small avivt@technion.ac.il}}%
}

\newtheorem{definition}{Definition}

\newcommand{\norm}[1]{\left\lVert #1 \right\rVert}

\def\mc#1{\mathcal{#1}}
\DeclareMathOperator{\E}{\mathbb{E}}

\definecolor{mygray}{gray}{0.6}

\newcommand{\probvoid}{GTSP-void}

\newcommand{\AT}[1]{\textcolor{magenta}{AT: #1}}
\newcommand{\KR}[1]{\textcolor{orange}{KR: #1}}

\date{}
\begin{document}

\begin{titlepage}
    \centering
    \mbox{ }\\[1cm]
    \LARGE
    {\bf Online Tool Selection with Learned Grasp Prediction Models}
    \vfill
    {\bfseries
        {\huge Technical Report}\\
        \vskip1cm
        {\Large 
            Date: \today
        }
    }    
    \vfill
    \small
    © OSARO Inc. All rights reserved. Materials may not be published, broadcast, rewritten, or redistributed without express written consent of OSARO Inc.
\end{titlepage}

\maketitle
\thispagestyle{empty}
\pagestyle{empty}

\begin{abstract}
Deep learning-based grasp prediction models have become an industry standard for robotic bin-picking systems. To maximize pick success, production environments are often equipped with several end-effector tools that can be swapped on-the-fly, based on the target object. Tool-change, however, takes time. Choosing the \textit{order} of grasps to perform, and corresponding tool-change actions, can improve system throughput; this is the topic of our work. The main challenge in planning tool change is \textit{uncertainty} -- we typically cannot see objects in the bin that are currently occluded. Inspired by queuing and admission control problems, we model the problem as a Markov Decision Process (MDP), where the goal is to maximize expected throughput, and we pursue an approximate solution based on model predictive control, where at each time step we plan based only on the currently visible objects. Special to our method is the idea of \textit{void zones}, which are geometrical boundaries in which an unknown object will be present, and therefore cannot be accounted for during planning. Our planning problem can be solved using integer linear programming (ILP). However, we find that an approximate solution based on sparse tree search yields near optimal performance at a fraction of the time.
Another question that we explore is how to measure the performance of tool-change planning: we find that throughput alone can fail to capture delicate and smooth behavior, and propose a principled alternative. Finally, we demonstrate our algorithms on both synthetic and real world bin picking tasks.

% A robotic bin-picking system in production environments requires
% grasping of a diverse range of novel objects randomly packed
% in a bin. Such systems must be capable of coping with objects
% of different weights, material types, shapes, and surface geometry. As a result
% robotic cells are often equipped with a tool selection mechanism
% that allows the robot to change the end-effector on the fly 
% mid process. Since tool change can take a significant amount of time, 
% it is important to optimize the swapping decisions such that the 
% total picking rate is maximized. In this paper we present a 
% planning framework for optimizing tool selection 
% by directly planning in the space of grasp proposals generated by pre-trained end-effector grasp prediction models. We introduce {\em Grasp Tool Selection Problem} (GTSP) which is based off a Markov Decision Process (MDP) formulation of the tool selection problem and present a model predictive control  (MPC) framework for solving it. As a secondary contribution we
% also present an integer linear program (ILP) adaptation of 
% GTSP by representing the system dynamics in terms  of set of linear constraints. We present results in a set of real world bin picking experiments
% benchamrking the performance of these approaches. 
\end{abstract}

%===============================================================================

\section{Introduction}
\label{sec:introduction}
\begin{figure*}
\centering
   \begin{subfigure}{0.13\textwidth}
   \centering
   \includegraphics[width=\linewidth]{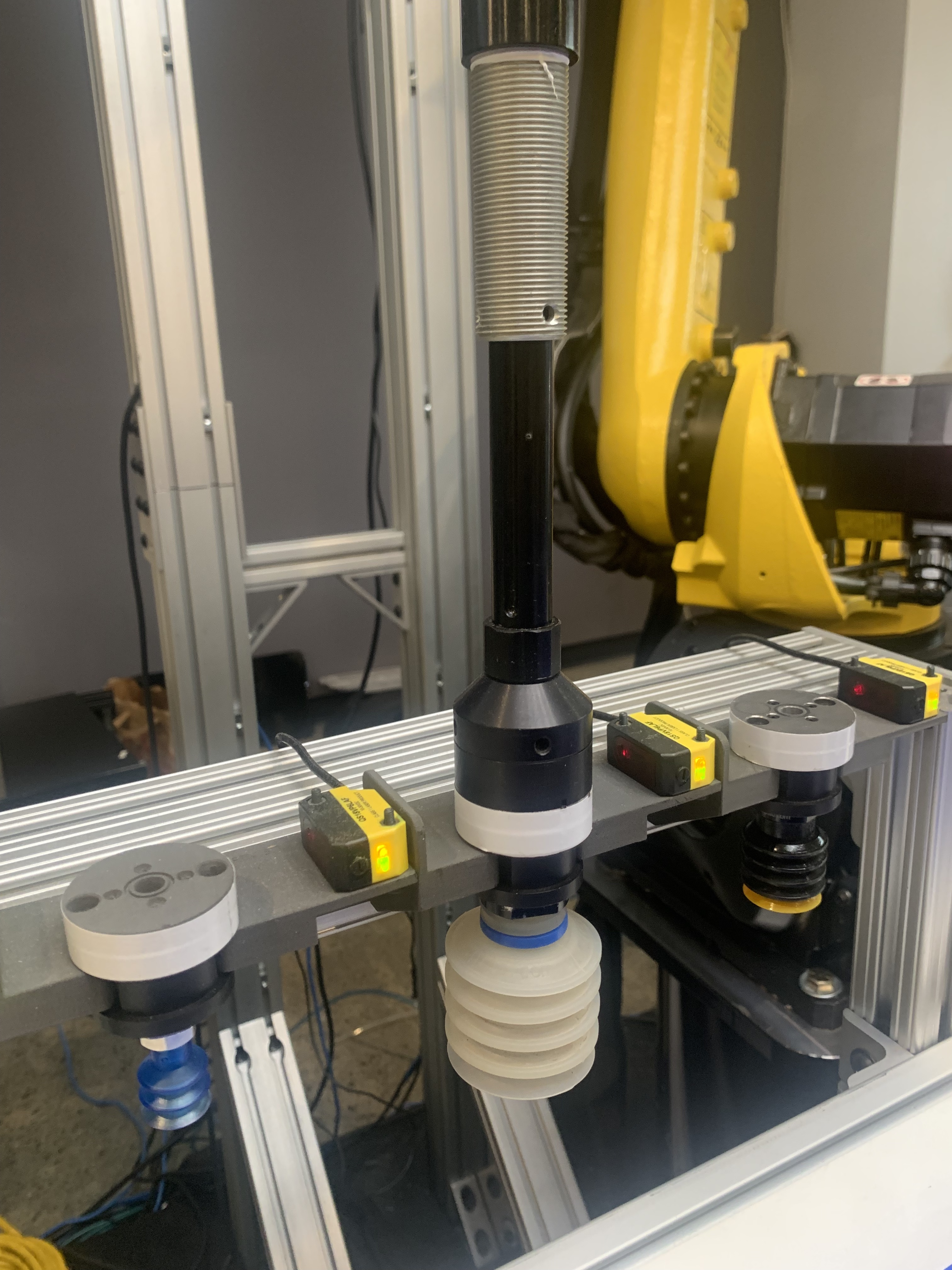}
   \caption{}
   \label{fig:f1} 
\end{subfigure}\quad
\begin{subfigure}{0.25\textwidth}
   \centering
   \includegraphics[width=\linewidth]{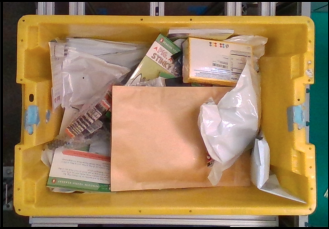}
   \caption{}
   \label{fig:f2}
\end{subfigure}
\begin{subfigure}{0.25\textwidth}
   \centering
   \includegraphics[width=\linewidth]{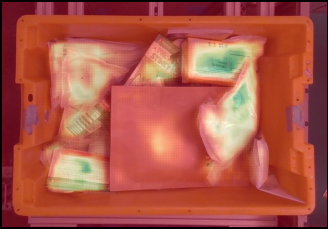}
   \caption{}
   \label{fig:f3}
\end{subfigure}
\begin{subfigure}{0.25\textwidth}
   \centering
   \includegraphics[width=\linewidth]{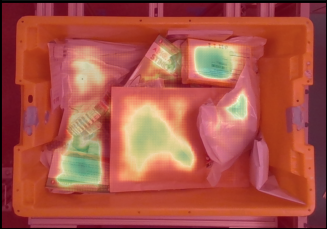}
   \caption{}
   \label{fig:f4}
\end{subfigure}
\centering
\caption{(a) Tool-changer station hosting different vacuum end-effectors of various sizes. The current  selection is already mounted on the robot; (b) Example of a bin containing an assortment of various objects; (c) and (d)  depict grasp score heat maps for a $30$mm and a $50$mm vacuum suction cups respectively (green spectrum denotes higher grasp scores). Note that the grasp prediction model for vacuum suction cup $50$mm generally places more probability mass over the larger objects compared to the $30$mm vacuum suction cup.}
\label{fig:tc-example}
% \vspace{-1.5em}
\end{figure*}

% \vspace*{-0.65cm}
Automated bin picking has gained considerable attention from manufacturing, e-commerce order fulfillment, and warehouse automation. The problem generally involves grasping of a diverse set of novel objects, which are often packed  randomly inside a bin (Figure~\ref{fig:tc-example}(b)). A common model-free bin picking approach is based on learning \textit{grasp prediction models} -- deep neural networks that map an image of the bin to success probabilities for different grasps~\cite{zeng2018robotic,lenz2015deep} (see Figures~\ref{fig:tc-example}(c) and \ref{fig:tc-example}(d) for examples of learned grasp prediction models for two different vacuum suctions with diameters $30$mm and $50$mm respectively).
% \AT{Talk about grasp prediction models - is there some formal reference?}
% Note that objects have different attributes in terms of shape, weight, texture, material type, and surface geometry. 
% \KR{Secdtion 2.2 in https://journals.sagepub.com/doi/pdf/10.1177/0278364919868017 discusses a range of previous work. Also, the original google robot farm paper (by Levine) has some connection.}

In order to handle a diverse range of objects, robotic cells are often equipped with a tool changer mechanism (Figure~\ref{fig:tc-example}(a)), allowing the robot to select a new end-effector from a set of available end-effectors (e.g., vacuum end-effectors varying in size, antipodal end-effectors) and swap it with the current one automatically in real time
(for example $50$mm vacuum end-effector grasp prediction model depicted in Figure~\ref{fig:tc-example}(d) places more probability mass over the larger objects, while $30$mm vacuum end-effector grasp prediction model model depicted in Figure~\ref{fig:tc-example}(d) places more probability mass over the smaller objects for the example bin image given in Figure~\ref{fig:tc-example}(b)). Choosing the right tool for each object can increase the pick success, potentially leading to improved throughput.
% \begin{figure}[H]
%     \centering
%     \includegraphics[width=0.36\linewidth]{figures/mixed-tote-1.jpg}
%     \hspace{0.1in}
%     \includegraphics[width=0.40\linewidth]{figures/tc-station.png}
%     \caption{(Left) Example of a bin containing a set  of randomly packed objects; (Right) Tool-changer station hosting three different vacuum end-effectors of various sizes. The current  selection is already mounted on the robot.}
% \label{fig:tc-example}
% \end{figure}
% \vspace{-0.5mm}
% Figure~\ref{fig:tc-example}(Right) shows an example of a tool-changer hardware that hosts a number of vacuum end-effectors varying in size (in terms of the suction cup diameter) and the strength of the suction airflow. While smaller vacuum end-effectors (in diameter) are naturally a better fit for tiny surfaces, larger vacuum end-effectors are proven to grasp more robustly on large surfaces and heavier objects. For every pick, the robot needs to make a decision on which end-effector to use next, and possibly swap the current one based on the decision. 
Tool changing, however, comes at a cost of cycle time: navigating the end-effector to the tool changing station, and performing the swap. \footnote{Mounting several tools on the same end effector, while possible~\cite{zeng2018robotic}, can be difficult, as, for example, different vacuum suction cups would require multiple hosing. A tool change station is a more scalable approach.} 
% in practice they might result in other challenges in terms of cable routing which would potentially cause interference with other sensors (e.g., force sensor).

In common picking tasks, the agent is free to choose the \textit{order} of objects to pick, and respectively, the order of tool changes. Thus, by carefully planning the picking order, we can potentially improve picking efficiency, by, e.g., using the same tool repeatedly for several objects. This is our main objective in this work.
% and also which tool to use, we hypothesize that \textit{planning} a sequence of picks and toll change actions can improve the \textit{overall} performance of the system.
% Thus, we need efficient algorithms that optimize the throughput while minimizing the tool changing costs. 
Optimizing tool selection, however, is challenging due to several reasons. Typically, some objects in the bin are occluded, and even objects that are currently visible may move unexpectedly due to grasp attempts of nearby objects, affecting their optimal tool selection in the future. Furthermore, even if the objects positions were known in advance, the complexity of computing the optimal picking order scales exponentially with the planning horizon, and is effectively intractable for real-time operation. Our goal in this work is to develop a scalable, well-performing, and fast method for approximately optimal tool selection. 

% \AT{The challenges are not clear, and its not clear how our approach addresses them}the combinatorial space of bin configurations (in terms of diversity of the objects randomly packed in a bin), high dimensional visual sensory inputs, and also due to the inherent uncertainties in sensing and control. 

We present a general formulation of this stochastic decision making problem, which we term {\em Grasp Tool Selection Problem} (GTSP), as a Markov Decision Process (MDP) (Section~\ref{sec:problem-formulation}). In practice, solving this MDP is difficult due to its large state space and difficult-to-estimate transition dynamics. To address this, we introduce an approximation of the problem where we: (1) replace the discounted horizon problem with a receding horizon Model Predictive Control (MPC); (2) In the inner planning loop of the MPC component, we replace the stochastic optimization with an approximate deterministic problem that does not require the complete knowledge of the true transition dynamics. We show that this deterministic problem is an instance of integer linear programming (ILP), which can be solved using off-the-shelf software packages. However, we further show that an approximate solution method based on a sparse tree search improves the planning speed by orders of magnitude, with a negligible effect on the solution quality, and is fast enough to run in real time.

% This problem formulation, which we term {\em Grasp Tool Selection Problem} (GTSP) is the main contribution of this work (Section~\ref{sec:gtsp}). To solve GTSP, we propose a sparse tree search implementation of MPC that leverages individual pre-trained end-effector grasp prediction models. As a secondary contribution we also present an integer linear program (ILP) formulation of GTSP by mapping it to the problem of minimum cost elementary shortest path subject to the constraints introduced by the approximate transition dynamics in GTSP (Section~\ref{sec:ilp}). Another novel contribution of this work is the derivation of a set of metrics for benchmarking tool selection algorithms
% which will be introduced in Section~\ref{sec:experiments}.

Our approach decouples grasp prediction learning from tool selection planning -- we only require access to a set of pre-trained grasp prediction models for each individual end-effector. Thus, our method can be applied ad hoc in production environments whenever such models are available. In our experiments, on both synthetic and real-world environments, we demonstrate that our planning method significantly improves system throughput when compared to heuristic selection methods. Another novel contribution of this work is the derivation of a set of metrics for benchmarking tool selection algorithms, based on practical considerations relevant to the bin picking problem.
% which will be introduced in Section~\ref{sec:experiments}.

% \AT{This seems like a different aspect, not related to MPC vs ILP perhaps discuss in a new sentence, without the number (3)}(3) we also assume that we have access to a set of pre-trained individual end-effector grasp prediction models and hence decouple grasp prediction learning from planning in this problem (in Section~\ref{sec:experiments} we briefly describe our proprietary pre-trained end-effector models that we used in our experiments, however please note that learning grasp prediction models is not the primary focus of this work). 

% \vspace{-0.3in}
\subsection*{Related Work} 
There is extensive literature on learning grasp prediction models~\cite{levine2016learning,redmon2015realtime,Mahlereaau4984}. To the best of our knowledge, few previous studies considered tool change optimization.
% there have not been many previous work related to this problem. 
The closest related work is~\cite{Mahlereaau4984} in the context of controlling
an ambidextrous robot for bin picking problem. That work focused on scaling the learning of the grasp prediction models, and not on the tool selection problem. In their approach, the best tool is selected greedily based on the grasp prediction scores generated by the end-effector grasp prediction models. 
Tackling problems with uncertain transitions by replanning using deterministic models is a common approach in planning~\cite{russell1995artificial} and robotics, where it is commonly referred to as model predictive control~\cite{camachoMPC2013,tamar2016}. To our knowledge, this work is the first application of this idea to tool change optimization with learned grasp models. 
% , also known as receding  horizon control, is a widely used model-based control method for problems with complex dynamics, with  a variety of applications in robotics and control~\cite{camachoMPC2013,tamar2016} and also 
Several studies focused on planning with deep visual predictors~\cite{finn2016,ebert2017selfsupervised,ebert2018visual,xie2019improvisation},
where a deep visual predictive model is learned and combined with MPC. Our work differs from these
approaches in that we perform planning directly in grasp proposal 
space, based on our void zone approximation.

\section{Grasp Tool Selection Problem (GTSP) Formulation}
\label{sec:problem-formulation}
%\vspace{-.5em}
We assume a planar workspace $\mc{I}$, discretized into a grid of $W \times H$ points. For example, $\mc{I}$ could be the bin image, as in Figure~\ref{fig:tc-example}(b), where every pixel maps to a potential grasp point in the robot frame. A grasp proposal evaluates the probability of succeeding in grasping at a particular point. Formally, a grasp proposal is a tuple $\omega \coloneqq \{\mc{E}, u, \rho\}$, where $\mc{E}$ is an end-effector, $u \in \mc{I}$ is a position to perform a grasp (e.g., a pixel in  the image), and $0 \leq \rho \leq 1$ is the probability of a successful grasp when the end-effector $\mc{E}$ is used to perform a grasp on position $u$. We also use the notations $\omega^\mc{E}, \omega^u$, and $\omega^\rho$ when referring to individual elements of a grasp proposal $\omega$.

A grasp prediction model $\Gamma_{\mc{E}}: \mc{I} \rightarrow \{\omega_i\}_{i=1}^{W \times H}$ gives a set of grasp proposals for an input image $\mc{I}$ and end-effector $\mc{E}$. In practice, only a small subset of grasp proposals yield good grasps. 
% (intuitively, in the order of the number of objects inside the bin). 
Thus, without loss of generality, we denote $\Gamma^k_{\mc{E}}(\mc{I}) = \{\omega_i\}_{i=1}^k$, limiting the model only to the $k$ best grasps (in terms of grasp success probability). 
Given a set of $n$ end-effector grasp proposal models $\{\Gamma_i\}_{i=1}^{n}$\footnote{For simplifying notations we interchangeably use $\Gamma_i$ to refer to an end-effectors grasp proposal model $\Gamma_{\mc{E}_i}$.}, we define the {\em grasp plan space} $\Omega \coloneqq \cup_{i=1}^{n} \Gamma^{ k}_i(\mc{I})$, which denotes the space of all plannable grasps. We will further denote by $\Omega_t$ the grasp plan space that is available at time $t$ of the task.
% \coloneqq \cup_{i=1}^{n} \Gamma^{t, k}_i(\mc{I})$

\paragraph{Grasp Tool Selection Problem (GTSP):} 
% We desire to optimize the throughput (e.g., number of successful grasps per hour) while minimizing the tool changing costs (e.g., minimizing tool changing frequency) in a long run. Therefore, 
We model the problem as a Markov Decision Process (MDP \cite{bertsekas2012dynamic}) $\mc{M} \equiv \langle \mc{S}, \mc{A}, \mc{R}, \mc{T}\rangle$ defined as follows: $\mc{S}$ is a set of states, where each $s = \langle \Omega, \mc{E}\rangle \in \mc{S}$ consists of the current grasp plan space and the current end-effector on the robot.
% with $s_t \in \mc{S}$ denoting the state at time $t$. We define $s_t \coloneqq \langle \Omega_t, \mc{E}_t\rangle$ which is a tuple consisting of the current grasp plan space and the end-effector on the robot at time $t$. 
We denote by $s^\Omega$ and $s^\mc{E}$ the individual elements of state $s$. 
The action space $\mc{A}=\Omega$ is the set of all plannable grasp proposals. 
% The action space $\mc{A}_t$ is defined as a set of all plannable grasp proposals at time $t$, in other words $\mc{A}_t \coloneqq \Omega_{t}$. 
The reward balances pick success and tool change cost, and we chose it as:
\vspace{-0.5em}
\begin{equation}
    {\mc{R}}(s, \omega) = \hspace{0.01in} \omega^{\rho} + c \hspace{0.02in} \mathbf{1}(s^{\mc{E}} \neq \omega^{\mc{E}}),
    \label{eq:mpc-reward}
\end{equation}
% $\mc{R}(s_t, \omega_t)$ is a reward function which  should reflect both grasp execution outcome (positive reward if grasp $\omega_t$ was successful, else $0$) and tool swapping costs (negative reward  if $\mc{E}_t \neq \omega^{\mc{E}}_t$). Throughout this paper we use a weighted sum of the grasp success prediction scores provided by the end-effector grasp prediction models, and the costs of tool change:
% \begin{equation}
%     {\mc{R}}(s_t, \omega_t) = \hspace{0.01in} \omega^{\rho}_t + c \hspace{0.02in} \mathbf{1}{(\mc{E}_{t} \neq \omega^{\mc{E}}_t)}
%     \label{eq:mpc-reward}
% \end{equation}
% \AT{$\mc{R}(s_t, \omega_t) = \omega^{\rho}_t + c \mathbf{1}(s_t^{\mc{E}} \neq \omega^{\mc{E}}_t)$}
where $\omega^{\rho}$ is the grasp success score, $c<0$ reflects a negative reward for tool changing, and $\mathbf{1}(\cdot)$ is the indicator function\footnote{Please note that this is only one way of designing a reward function for this problem. In general more interesting types of reward function could be crafted by reflecting additional problem specific requirements. Examples of which are including the costs of robot motion in terms of the distance travelled between consecutive grasp proposals, or the proximity of the end-effector with respect to the tool changer station.}. 
% where $\omega^{\rho}_t$ is the grasp success score associated with the grasp proposal $\omega_t$ at time $t$, $c<0$ reflects a negative reward for tool changing, and $\mathbf{1}(.)$ is the indicator function. 
The state transition function $\mc{T}(s_t, \omega_t, s_{t+1}) \rightarrow [0, 1]$ gives the probability of observing the next state $s_{t+1}$ after executing grasp proposal $\omega_t$ in state $s_t$. As a result of performing a grasp, and depending on the grasp outcome, an object is removed from the bin and some other object randomly appears at the position of the executed grasp and new graspable positions will be exposed. The optimal policy $\pi^*:\mc{S}\to\Omega$ is defined as: $\pi^* = \operatorname*{argmax}_\pi \E_{\pi}\left[ \sum_{t=0}^{\infty} \gamma^t \mc{R}(s_t, \omega_t) | \omega_t\sim\pi(s_t)\right]$.
% , and maximizes the expected long term return.
% by either deciding to keep the current end-effector, or selecting a different one for executing the next grasp.     

%===============================================================================

\vspace{-.5em}
\section{Approximate GTSP Solution}
\label{sec:gtsp}
%\vspace{-.5em}
Solving GTSP is difficult for following reasons:
\textbf{(1) Prediction:} we do not know the true state transitions, as they capture the future grasps that will be made possible after a grasp attempt, which effectively requires a prediction of objects in the bin that are not directly visible (see, e.g., Figure~\ref{fig:tc-example}(b)). Although several studies investigated learning visual predictive models (see~\cite{finn2016,ebert2017selfsupervised,ebert2018visual,xie2019improvisation}), learning such models in production environments with a high variability of objects is not yet practical. \textbf{(2) Optimization:} even if the state transitions were known, the resulting MDP would have an intractable number of states, precluding the use of standard MDP optimization methods.
% (a GTSP with $m$ end-effectors and $n$ sampled grasp proposals per end effector would yield an upper bound of $O(m \hspace{0.01in}2^{(mn)})$\AT{why 2?} states). 

    % for a moderate number of possible objects and end effectors, the resulting state space of the MDP is enormous.

% n accurate visual scene change prediction. Although there have been several studies aimed at learning visual predictive models (see~\cite{finn2016,ebert2017selfsupervised,ebert2018visual,xie2019improvisation}), learning such models in production environments with high variability in the objects is not yet practical. For example, when an object is grasped, it is hard to tell in advance whether the grasp attempt will move adjacent objects, and it is also difficult to predict what object lies underneath the grasped object (see Figure~\ref{fig:tc-example}(Left)).

% First, even for a moderate number of possible objects and end effectors, the resulting state space of the MDP is enormous. Second,  

We address (1) by replacing the stochastic optimization with an approximate deterministic problem that does not require the complete knowledge of the true transitions, based on the idea of \textit{void zones}.
We address (2) by replacing the infinite horizon discounted horizon problem with an online solver, which at each time step chooses the grasp that is optimal for the next several time steps; we term this part the model predictive control (MPC) component. We propose two computational methods for solving the short horizon problem in the inner MPC loop, either accurately, based on integer linear programming (ILP) outlined in Section~\ref{sec:ilp}, or approximately, using a sparse tree search method (STS) outlined in Section~\ref{sec:sts}. 

Algorithm~\ref{alg:ts-mdp} outlines the generic Model Predictive Control (MPC) for solving the GTSP. At every step, the current observations (e.g., bin image $\mc{I}$) is fed to the set of pre-trained end-effector models to obtain the plan space $\Omega_t$. Next, a \probvoid-Solver is called over the current state $s_t \gets \langle \Omega_t, \mc{E}_t\rangle$ which will return an optimized plan $\omega$, and finally the first step (i.e., grasp proposal) of the plan is executed. We have already described the ILP solver in the paper and will outline the Sparse Tree Search (STS) solver in the next section.
\begin{algorithm}%[hbt!]
% \small
	\caption{Model Predictive Control (MPC)}\label{alg:mpc}
	\begin{algorithmic}[1]
	\State {\bf Inputs}: Set of end-effector models $\{\Gamma_i\}_{i=1}^n$, maximum number of best grasp proposals per end-effector model to include $m$, receding horizon $H$, tool changing costs $c < 0$, void radius $l$
		\For {$t=1, 2, \ldots$}
		    \State Observe the current bin image $\mc{I}_t$, and  current end-effector $\mc{E}_t$
		    \State Construct the current plan space: $\Omega_t \gets \cup_{i=1}^{n} \label{alg:plan-space} \Gamma^m_i(\mc{I}_t)$
		    \State $s_t \gets \langle \Omega_t, \mc{E}_t\rangle$  
		    \State $(\omega, v_{\omega}) \gets$ \Call{\probvoid-Solver}{$s_t, H, c, l$} \label{alg:mpc-inner-function} \Comment{\textcolor{gray}{ILP, or Sparse-Tree-Search (STS) }}
			\State Execute the grasp proposal $\omega$
		\EndFor
	\end{algorithmic} 
\label{alg:ts-mdp}
\end{algorithm}

\subsection{Approximate Prediction using Void Zones}
\label{sec:void_zones}
% In GTSP, we also 
We seek to replace the stochastic (and unknown) transitions in GTSP 
% original tool selection MDP 
by deterministic dynamics, such that the solution of the deterministic problem will yield a reasonable approximation of the true optimal controller. Our approach is based on the idea of a \textit{void zone} -- not allow a grasp that is in very close proximity to any previous grasp, as movement of objects in the bin resulting from the previous grasp attempt could render a future grasp in its close vicinity impossible. 
% Thus, we only plan grasp sequences that are guaranteed to be feasible, regardless of the unknown transitions. While this may result in a sub-optimal plan (as some actions are not allowed), it does not require any prediction of the scene change. 

We motivate void zones with the following working hypothesis: \textit{As long as the objects are sufficiently small, when a grasp is attempted, 
% it can only impact a local neighborhood 
% of the grasped object in the bin. Thus, 
the set of grasp proposals that are sufficiently distant 
from the attempted grasp position will remain valid in the next state. }

% \begin{center}
% \begin{minipage}{0.95\textwidth}
% \textit{As long as the objects are sufficiently small, when a grasp is attempted, it can only impact a local neighborhood 
% of the grasped object in the bin. Thus, the set of grasp proposals that are sufficiently distant 
% from the attempted grasp position will remain valid in the next state. }
% \end{minipage}
% \end{center}

\begin{figure}
    % \vspace{-1em}
    \centering
    \includegraphics[width=1.5in]{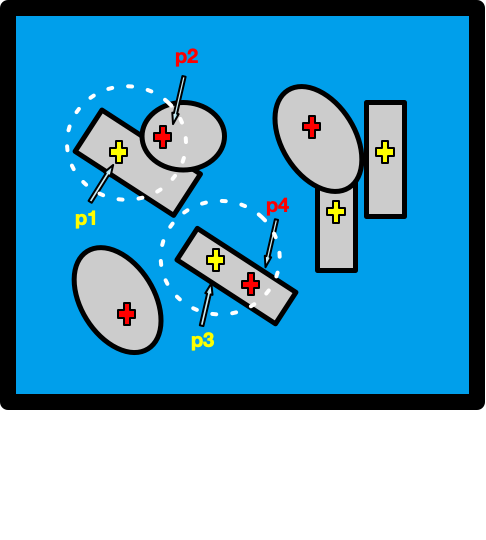} \hspace{0.1in}
    \includegraphics[width=1.5in]{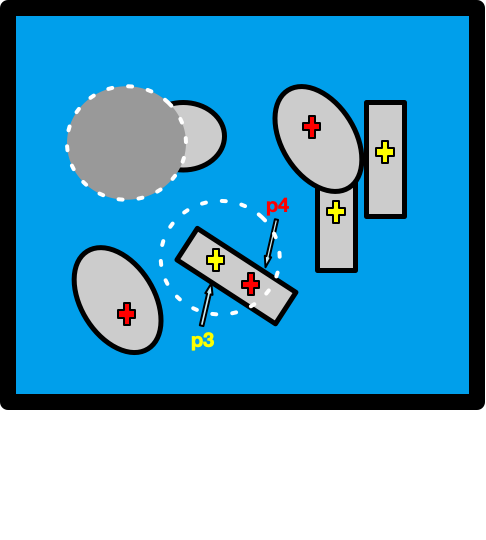}
    % \vspace{-0.3in}
    \caption{(Left) examples of cases where performing a grasp proposal would invalidate some other grasp proposals ($p_1$ invalidates $p_2$, and $p_3$ invalidates $p_4$); (Right) updated grasp plan space for the next horizon step by voiding the impacted grasp proposal (in this case voiding $p_2$ as a result of committing to the grasp proposal $p_1$).}
    % \caption{A toy example of a bin picking problem with two end-effectors. The grasp proposals are color coded for each end-effector: (Left) examples of cases where performing a grasp proposal would invalidate some other grasp proposals ($p_1$ invalidates $p_2$, and $p_3$ invalidates $p_4$). The impact area for each case is approximately represented by the dashed white circle; (Right) We update the grasp plan space for the next horizon step by voiding the impacted grasp proposal (in this case voiding $p_2$ as a result of committing to the grasp proposal $p_1$). The gray area represents the area in which all potential grasp proposals are voided for the next horizon step.}
    \label{fig:tc-dynamics}
    % \vspace{-2em}
\end{figure}This observation is illustrated in Figure~\ref{fig:tc-dynamics}(Left), for a bin picking problem with two end-effectors. The grasp proposals are color coded for each end-effector. In some cases, grasp proposals lie over different objects where one object might be partially occluding the other one (e.g., $p_1$ and $p_2$). In other cases, two or more grasp proposals might lie on the same object. In either case, performing one of the grasp proposals will invalidate some other grasp proposal and hence those proposals should not be available to the planner in the next steps. 

We define the void zone based on the Euclidean distance, as follows:
\begin{definition}[{\em l-separation}]\label{def:l-seperated} 
  Let $d_{i,j}= \norm{\omega^u_i - \omega^u_j}$ denote the Euclidean distance on the plane between grasp proposals $\omega_i$ and $\omega_j$. A pair of grasp proposals $\langle \omega_i, \omega_j\rangle$ is called {\em $l$-separated} if $d_{i,j} > l$. We refer to $l$ as void radius and use the notation $\Psi_l(\omega)$ to  refer to a set of grasp proposals which are l-separated from $\omega$. Note that by definition $\omega \not\in \Psi_l(\omega)$.
  \label{def:l-sep}
\end{definition}

% The rest of the  grasp proposals sufficiently distant from them remain valid for the next plan step. 
% Identifying grasp proposals that may invalidate future grasps is challenging, since
% detecting object boundaries is difficult.
% At the same time, without object boundaries, we cannot know if two grasp points belong to the same object or not, or whether performing one would potentially render the other one invalid. 

% Identifying  grasp  proposals that  may  invalidate future grasps  is  challenging because of the complicated dynamics of state transitions — in addition to items being removed from the bin, the picking action may shift or cause surrounding items to fall over. Instead, we propose to simplify the problem by addition of a {\em void zone} to each grasp proposal, such that if we choose a grasp proposal, we are not allowed to plan additional grasps within its void zone. 

Based on the above definition, we can formally define deterministic dynamics in GTSP, which we will henceforth refer to as \textit{\probvoid}. At state $s_t=\langle \Omega_t, \mc{E}_t\rangle$, taking action $\omega_t$ results in a next state,
\begin{align}
\small
    s_{t+1} =& \langle \Omega_{t+1}, \omega^{\mc{E}}_t\rangle ,
    \nonumber
    \\
    % \qquad
    \Omega_{t+1} =& \{\omega \hspace{0.05in} | \hspace{0.05in} \omega \in  \Omega_t \land \omega \in \Psi_l(\omega_t)\}
    \label{eq:mdp-dynamics}
\end{align}
% for current state $s_t=\langle \Omega_t, \mc{E}_t\rangle$ and  current action $\omega_t$. 
That is, the end-effector in the next state is as chosen by $\omega_t$, and the grasp plan space is updated to exclude all grasp proposals within the void zone.

As shown in Figure \ref{fig:tc-dynamics}, by setting the void zone large enough, we can safely ignore the local changes as a result of executing a grasp. Obviously, using void zones comes at some cost of sub-optimality -- as we ignore possible future grasps inside the void zones. To mitigate this cost, we propose a model predictive control (MPC) approach.
% , as specified in Algorithm~\ref{alg:ts-mdp}. 
At every step, the current observation (i.e., bin image $\mc{I}$) is fed to the set of pre-trained end-effector models to obtain the plan space $\Omega_t$. Next, we solve the corresponding \probvoid\ problem with some fixed horizon $H$, and the first step of the plan $\omega$ is executed.

% At every step, the current observation (i.e., bin image $\mc{I}$) is fed to the set of pre-trained end-effector models to obtain the plan space $\Omega_t$ (line \ref{alg:plan-space}). Next, we solve the corresponding \probvoid\ problem with some fixed horizon $H$ (line~\ref{alg:mpc-inner-function}), and the first step of the plan $\omega$ is executed. 

% routine (line~\ref{alg:mpc-inner-function}) in the MPC controller 
Replanning at every step allows our method to adapt to the real transitions observed in the bin. Next, we propose two methods for solving the inner \probvoid\ optimiztion problem within our MPC. 

\subsection{Approximate Optimization using Integer Linear Programming}
\label{sec:ilp}
% The MPC algorithm presented in Section~\ref{sec:mpc} is brute-force and has an exponential complexity $O(|\Omega|^H)$ in grasp plan space. In practice we are forced to work with small horizons $H$ and expand the MPC tree search over a small subset of grasp proposals at every (see line \ref{alg:mpc-subsample} in Algorithm~\ref{alg:mpc-func}) which results in sub-optimal solutions. 

% In this section we present an integer linear program (ILP) formulation of the MPC algorithm which can be used in Algorithm~\ref{alg:mpc} (see line~\ref{alg:mpc-inner-function}). 
In this section we show that the \probvoid\ problem can be formulated as an integer linear program (ILP).
% Although the ILP formulation requires an exponential number of constraints, we can leverage
% highly optimized solvers
% \footnote{In all of our experiments we used {\em Gurobi} optimizer~\cite{gurobi}.} 
% to solve \probvoid\ optimally in the inner MPC optimization step. 
To motivate this approach, note 
that \probvoid\ with horizon $H$ seeks to find a trajectory of $H$ $l$-separated grasp proposals in $\Omega_t$ with the highest accumulated return. This motivates us to think of the problem as a walk over a directed graph generating an {\em elementary path}\footnote{In an elementary path on a graph all nodes are distinct. } of length $H$ of $l$-separated grasp proposals with the highest return, where the nodes of the graph are the grasp proposals in the current state, $s^{\Omega}_t$, and the directed edges represent the order at which the grasp proposals are executed. 
% Each directed edge is also associated with a reward of executing the pair of grasp proposals according to the edge direction (see Equation~\ref{eq:mpc-reward} in  Section~\ref{sec:problem-formulation}). Note the graph topology dynamically changes every time we execute a grasp proposal due to the $l$-separation requirement (if we select a grasp proposal, we will  have to remove all the graph nodes which are not $l$-separated from the current selection) and will depend on the selection order. Thus, for every pair of grasp proposals (or nodes in the graph), we need two directed edges, each connecting one to the other. A major contribution of this work is how to model such a dynamic process in terms of the $l$-separation requirement represented as a set of linear constraints. 
Our formulation is mainly inspired by the ILP formulation of the well known {\em Travelling Salesman Problem} (TSP)~\cite{danzig1954} with the following changes: (1) the main objective is modified to finding an elementary path of length $h$ with maximal return, anywhere on the graph (as opposed to the conventional {\em tour} definition in TSP); (2) addition of the $l$-separation constraints to enforce voiding; (3) a modification of {\em Miller-Tucker-Zemlin} sub-tour elimination technique \cite{mtz1960} for ensuring the path does not contain any sub-tour.

\begin{comment}
directed graph where the nodes are the current grasp proposals and directed edges represent the step reward of executing a pair of grasps in that order (note that for every pair of grasp proposals we have two directed edges, each associated with a reward of performing them in that order). Based on this view, we can think of MPC objective as the problem of finding an elementary path of length $H$ of $l$-separated grasp proposals with maximum return accumulated along that path. We now show that the $l$-separation requirement \probvoid\ can be represented by a set of linear constraints.
\end{comment}

\begin{algorithm}
% \renewcommand{\thealgorithm}{}
% \floatname{algorithm}{}
\caption{Integer Linear Program (ILP)}\label{alg:ilp}
\begin{algorithmic}[1]
% \Function{ILP}{$s_t, H, c, l$}: current state $s_t = \langle \Omega, \mc{E} \rangle$; horizon $H$, cost $c < 0$, void radius $l$
\State Objective: $v^* = \underset{\omega_{ij}}{\text{max}} \sum_{w_{ij}} r_{ij} \omega_{ij}  \hspace{0.1in} \mbox{s.t.}$ \label{alg:ilp-obj}
\State $\omega_{ij} \in \{0, 1\}, \hspace{0.05in} \forall{i, j \in \mc{V}}$ \label{alg:ilp-binary} \Comment{\textcolor{mygray}{binary variables representing edges in the solution}}
\State $\sum_{j \in \mc{V}\backslash\{s, e\}}^{} \omega_{sj} = 1$ \label{alg:ilp-source} \Comment{\textcolor{mygray}{constraint enforcing one outgoing edge from the initial node $s$}}
\State $\sum_{i \in \mc{V}\backslash\{s, e\}} \omega_{ie} = 1$ \label{alg:ilp-sink} \Comment{\textcolor{mygray}{constraint enforcing one incoming edge to the terminal node $e$}}
\State $\sum_{i, j \in \mc{V}\backslash\{s, e\}} \omega_{ij} = H$ \label{alg:ilp-horizon} \Comment{\textcolor{mygray}{constraint enforcing the trajectory length to be exactly $H$}}
\State $\sum_{j \in \mc{V}\backslash\{s, i, e\}} \omega_{ij}\hspace{-0.05in}=\hspace{-0.01in}\sum_{j \in \mc{V}\backslash\{s, i, e\}} \omega_{ji},  \hspace{0.05in} \forall i \in \mc{V}\backslash\{s, e\}$ \label{alg:ilp-flow-1}
\State $\sum_{j \in \mc{V}\backslash\{s, i, e\}} \omega_{ij} \leq 1$ \label{alg:ilp-flow-2} \Comment{\textcolor{mygray}{Steps $6$ and $7$ specify constraints enforcing the flow conservation}}
\State $\omega + \omega' \leq 1,\hspace{0.05in} \forall \omega \in \phi_i, \forall \omega' \in \phi_j,\hspace{0.05in} \mbox{if} \norm{\omega^u_i - \omega^u_j} \leq l, \forall i, j \in \mc{V}\backslash\{s, e\}$  \label{alg:ilp-void} \Comment{\textcolor{mygray}{$l$-separation constraints (see text)}}
\State  $u_i \in \{0, 1, 2, \hdots, |\mc{V}|\}, \hspace{0.05in} \forall{i \in \{1, \hdots, \mc{V}\}}$ \label{alg:ilp-mtz1}
\State $u_1 = 1$ and $u_{|\mc{V}|}=|\mc{V}|$, \quad
\label{alg:ilp-mtz2}
$2 \leq u_i \leq |\mc{V}|-1$, $\forall{i \in \{2, \hdots,  |\mc{V}|-1\}}$ \label{alg:ilp-mtz3} 
\State $u_i - u_j + 1 \leq (H+1)(1-\omega_{ij})$, $\forall{i, j \in \{2, \hdots,  |\mc{V}|-1\}}$ \label{alg:ilp-mtz4}  \Comment{\textcolor{mygray}{steps $9$-$11$ specify Miller-Tucker-Zemlin (MTZ) sub-tour elimination constraints}}
\State return $(\omega_i, v^*) \hspace{0.05in} \mbox{s.t.} \hspace{0.05in}  \omega_{si}=1$ 
% \EndFunction
\label{alg:ilp-return}
\end{algorithmic}
\end{algorithm}

Given the current state $s_t = \langle \Omega_t, \mc{E}_t\rangle$, we represent the grasp plan space as a graph $\mc{G} = \langle \mc{V}, E\rangle$ where the nodes of the graph are grasp proposals $\omega_i \in s^{\Omega}_t$ plus two auxiliary initial and terminal nodes
% source and sink nodes for the initial and terminal grasps
$\{s, e\}$: $\mc{V} = \{1, \hdots, |\Omega_{t}|\} \cup \{s, e\}$. We index the initial and terminal nodes by $1$ and $|\mc{V}|$, respectively. 
% These auxiliary nodes are for the initial and terminal grasps.
% will allow the algorithm to start the elementary path at any grasp proposal node and end at any other grasp proposal node). When indexing into $\mc{V}$, without a loss of generality, we assign the index $1$ to the source node, and assign the index $|\mc{V}|$ to the sink node. 
For any pair of $l$-separated grasp proposals $\omega_i$ and $\omega_j$ ($i, j \in \mc{V}\backslash\{s, e\}$) we add directed edges $\{e_{ij}, e_{ji}\} \in E$ with a reward $r_{ij} \coloneqq \rho_{\omega_i} + c \mathbf{1}{(\omega^{\mc{E}}_i \neq \omega^{\mc{E}}_j)}$ (cf. Equation~\ref{eq:mpc-reward}). For such pairs of grasp proposals we also add binary variables $\{\omega_{ij}, \omega_{ji}\}$ to ILP. We connect the initial node $s$ to the set of all grasp proposal nodes with reward defined as $r_{si} \coloneqq c \hspace{0.02in} \mathbf{1}{(\omega^{\mc{E}}_s \neq \omega^{\mc{E}}_i})$
and add binary variables $\omega_{si} \hspace{0.02in}(\forall i \in \mc{V}\backslash\{s, e\})$ to ILP. We also connect the set of all grasp proposal nodes to the terminal node $e$ with reward $r_{ie} \coloneqq \omega^{\rho}_i$,
and add corresponding binary variables $\omega_{ie}  \hspace{0.02in}(\forall i \in \mc{V}\backslash\{s, e\})$ to ILP. The optimization objective is  defined as maximization $v^* = \underset{\omega_{ij}}{\text{max}} \sum_{w_{ij}} r_{ij} \omega_{ij}$ subject to a set of constraint that enforce an {\em elementary path} of length $H$.

The complete ILP  formulation is outlined in Algorithm~\ref{alg:ilp}.  The constraints in lines \ref{alg:ilp-binary}-\ref{alg:ilp-flow-2} are similar to a standard TSP formulation~\cite{danzig1954}. Constraints on line~\ref{alg:ilp-binary} enforce binary constraints on $\omega_{i,j}$ variables where $\omega_{i,j} = 1$ selects the pair of grasp proposals $\omega_i$ and $\omega_j$ to be included in the solution, in that order. Constraint on line~\ref{alg:ilp-source} ensures only one outgoing connection exists from the auxiliary start node $s$ to grasp proposal nodes (marking the start of the path). Constraint on line~\ref{alg:ilp-sink} ensures only one incoming connection exists from grasp proposal nodes to the auxiliary sink node $e$ (marking the end of the path). Constraint on line~\ref{alg:ilp-horizon} enforces the length of the elementary path to be exactly $H$. Constraints on line~\ref{alg:ilp-flow-1} are flow conservation constraints, while the constraint on line~\ref{alg:ilp-flow-2} ensures that the outgoing degree of each node is at most one (standard flow conservation constraints in TSP (see~\cite{danzig1954})). 
Line~\ref{alg:ilp-void} defines the $l$-separation constraints. We denote by $\phi_i \coloneqq \{\omega_{j,k} | j=i \vee k=i\}$ the set of all incoming and outgoing edges of the node $i$.
% the set of all pairs of grasp proposals that involve the grasp proposal $i$ (i.e., the set of all incoming and outgoing edges of the node $i$). 
% Since all the grasp proposals are known in advance, we can iterate through every pair of grasp proposals and check if they are $l$-separated. If they are not, then 
For two nodes $i,j$ that are not $l$-separated, the constraint only allows for at most one element of $\phi_i,\phi_j$ to be included in the solution. The constraints on lines~\ref{alg:ilp-mtz1}-\ref{alg:ilp-mtz4} specify our adaptation of the {\em Miller-Tucker-Zemlin} sub-tour elimination technique \cite{mtz1960}. For each node in the graph (including the source and sink nodes) we add an integer variable $u_i$.
(with $u_1$ associated with the source node $s$ and $u_{|\mc{V}|}$ associated with the source node $e$). We enforce  
$u_1=1$ and $u_{|\mc{V}|}=|\mc{V}|$ to ensure the source node $s$ marks the start of the trajectory and the sink node $e$  marks the end of the trajectory (line~\ref{alg:ilp-mtz2}). The rest  of variables $u_i$ could take on values in between (line~\ref{alg:ilp-mtz3}). These constraints, together with 
line~\ref{alg:ilp-mtz4} induces an ordering of the grasp proposals which prevents sub-tours.

\subsection{Approximate Optimization using Sparse Tree Search}
\label{sec:sts}
%\vspace{-.5em}
\begin{comment}
We now present a simple alternative to ILP for approximately solving \probvoid\, 
based on a {\em sparse tree search}. Our approach, outlined in Algorithm~\ref{alg:mpc-func}, at every horizon\AT{what does at every horizon mean?} iteratively selects from a subset of grasp proposal (lines~\ref{alg:mpc-subsample} and \ref{alg:mpc-function-main-loop}) and recursively calculates a sub-plan rooted at that node for a maximum horizon of $H$ (line~\ref{alg:mpc-function-recursion}). 
It then accumulates the results of each recursion and computes the value of the sub plan (line~\ref{alg:mpc-value}). Finally, it returns the best sub-plan and its value among all the sub-plans calculated at that horizon. For selection of the subset of grasp proposals (line~\ref{alg:mpc-subsample}) we uniformly sample the next best grasp proposal per end-effector, although other sampling strategies are also possible. \AT{I think a simpler explanation would be that we're performing a tree search of depth $H$ over a reduced set of actions. Is that correct?}
\end{comment}
We now present a simple alternative to ILP for approximately solving \probvoid\, 
based on a {\em sparse tree search} (STS). Our approach, outlined in Algorithm~\ref{alg:mpc-func}, performs a tree search of depth $H$ where tree node expansion takes place  over a {\em sparse} subset of grasp proposals respecting the {\em l-separation} constraint (see Definition~\ref{def:l-sep} in Section~\ref{sec:void_zones}). At every search step a node is expanded using a sparse subset of available grasp proposals (line~\ref{alg:mpc-subsample}). In our approach we use the union of top $k$ grasp proposals per end-effector according to the grasp  proposal scores $\omega^{\rho}$. The parameter $k$ -- hereafter the {\em sparsity factor} -- determines the sparsity of the subset of grasp proposals for the tree search node expansion. While expanding over the set of all available grasp proposals at every node is possible and optimally solves the problem in theory, in practice it makes the planning significantly slow and hence is not suitable for real world production environments. The algorithm then recursively calculates a sub-plan rooted at that node for a receding horizon of $H$ (line~\ref{alg:mpc-function-recursion}), accumulates the results of each recursion and computes the value of the sub plan (line~\ref{alg:mpc-value}), and it returns the best sub-plan and its value among all the sub-plans calculated at that horizon.
% Our approach performs a tree search of depth $H$. where tree node expansion takes place  over a sparse subset of grasp proposals respecting the l-separation constraint. 
% At every search step a node is  
% expanded using a sparse subset of available grasp proposals, which we chose to be the top $k$ grasp proposals per end-effector according to the grasp  proposal scores $\omega^{\rho}$ (we only choose among grasp proposals that satisfy the the l-separation constraint). The parameter $k$ -- termed the {\em sparsity factor} -- therefore determines the sparsity of the grasp proposals subset. 

\begin{algorithm}
\small
\caption{Sparse Tree Search (STS)}\label{alg:mpc-func}
\begin{algorithmic}[1]
\Function{STS}{$s_t, H, c, l, k$}\Comment{current state $s_t = \langle \Omega_t, \Gamma_t \rangle$; horizon $H$, tool changing costs $c < 0$, void radius $l$, $k$ is the sparsity factor where it specifies the top $k$ grasp proposals per end-effector to be included for the search tree node expansion}
\If{$H = 0$} 
    \State return $(\_, 0)$
\EndIf 
\State $\Xi \leftarrow \emptyset$
\State Let $\hspace{0.02in} \Lambda^k_t \subset \Omega_t$ be the union of the top $k$ grasp proposals per end-effector (in terms of grasp score $\omega^{\rho}$) available in $\Omega_t$\label{alg:mpc-subsample}
\For {$\omega \in \Lambda^k_t$} \label{alg:mpc-function-main-loop}
\Comment{\textcolor{mygray}{$\omega = \langle \mc{E}, u, \rho\rangle$}}
    \State $s_{t+1} = \mc{T}_l(s_t, \omega)$ \label{alg:mpc-next-state} \Comment{\textcolor{mygray}{forward dynamics in Equation~\ref{eq:mdp-dynamics}}}
\State $(\omega^+, v_{\omega^+}) \gets$ \Call{STS}{$s_{t+1}, H-1, c, l, k$} \label{alg:mpc-function-recursion}
\State $v_{\omega} = \bar{\mc{R}}(s_t, \omega) + v_{\omega^+}$  \label{alg:mpc-value} \Comment{\textcolor{mygray}{reward function defined in~Equation\ref{eq:mpc-reward}}} \label{alg:mpc-function-val}
\State $\Xi \gets \Xi \cup (\omega, v_{\omega})$
\EndFor
\State \textbf{return} $\operatorname*{argmax}_{v_{\omega}} (\Xi)$
\State
\EndFunction
\end{algorithmic}
\end{algorithm}

%AT: just to make sure I understand this example - the top object in the stack needs the same tool as the other objects, right? Otherwise it won't make any difference as we need to remove the top object before attending to the bottom ones.

% \section{GTSP as an Integer Linear Program}
% \label{sec:ilp}
% \input{ilp}

%===============================================================================

% \section{Approximate GTSP optimization}
% \label{sec:gtsp}
% \input{gtsp}

%===============================================================================

\section{Experiments}
\label{sec:experiments}
We divide our presentation to an investigation on synthetic problems, aimed at quantifying our algorithmic choices, and a real robot study evaluating the performance of our method in practice.

\subsection{Synthetic Experiments}
\label{sec:synthetic-exp}
In the first set of experiments we conducted a comparative analysis of the two \probvoid\ solvers outlined in Sections~\ref{sec:ilp} and \ref{sec:sts}. Our goal is to answer the following question: How do the ILP and STS solvers compare in terms of the optimization quality and speed? 

% \begin{wrapfigure}{r}{0.4\textwidth}
%     \vspace{-2em}
%     \includegraphics[width=2.2in]{figures/synthetic-grasp-maps-2ee.png}
%     \caption{Synthetic end-effector grasp proposal model generation for two end-effectors generated over a fixed grid resolution $70$ $\times$ $110$.}
%     \label{fig:synthetic-grasp-map}
%     %\vspace{-2em}
% \end{wrapfigure} 

\begin{figure}
    \begin{center}
    \includegraphics[width=\textwidth]{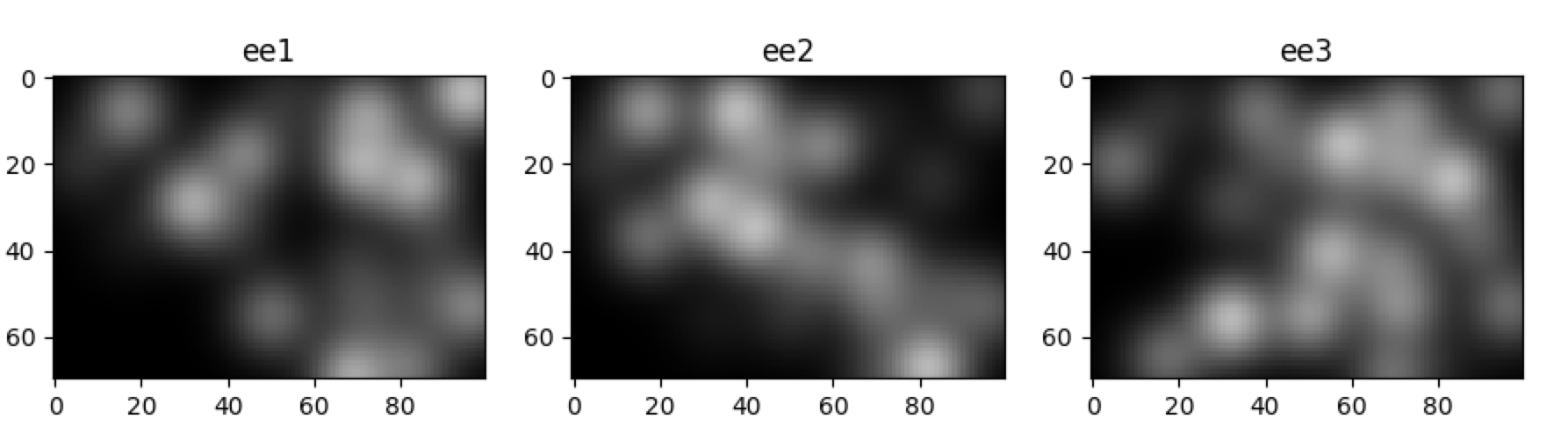}
    \caption{Synthetic end-effector grasp proposal model generation for three end-effectors generated over a fixed grid resolution $70 \times  110$.}
    \label{fig:synthetic-grasp-map}
    \end{center}
\end{figure} 

We crafted a synthetic tool selection problem generator as follows. 
% instances and use them to evaluate and compare the two solvers. 
A problem instance $\mc{T}$ is generated by first selecting the number of end-effectors and then, for each end-effector, we generate a random set of grasp proposals over a fixed grid resolution H$\times$W (we used H=$70$, W=$110$ in our experiments) \footnote{We generate grasp proposal sets directly, without requiring an image (cf. Section~\ref{sec:problem-formulation}).}
To generate realistic grasp proposals, we first choose $n=25$ random object positions, uniformly sampled on the grid.
% within a grid $70$$\times$$110$. 

Next, for each-end effector we generate random Gaussian kernels with randomized scale and standard deviation, centered on each object position. The resulting grasp proposal grid for each pixel gives a higher probability of success to pixels that are closer to an object center. 

Algorithm~\ref{alg:synthetic-gras-map} outlines the pseudocode for generating synthetic experiments. Function {\em GenerateGraspModel} generates a ransom grasp proposal model for a given resolution  $\langle H, W \rangle$ with higher grasp scores centered on a set of 
pixel positions $p^i = \langle p^i_x, p^i_y \rangle$ (mimicking object centers).This is simply done by defining randomly scaled normal distributions with means centered on positions $p^i$ and random standard deviations $\sigma_i$. Function {\em RunExperiment} describes the main loop for generating synthetic experiments. Given an ablation over experiment parameters (i.e., number of end-effectors $n_{ee}$, horizon $h$, sparsity factor $k$, void radius $l$, tool changing costs $c < 0$) and a number of episodes $N$ (for calculating the statistics), we first sample $m$ positions (mimicking object centers) and then generate  
$n_{ee}$ random grasp models using {\em GenerateGraspModel}. Next,
we run both ILP and STS solvers and collect performance results in  terms of plan value and plan time for each solver. Figure~\ref{fig:synthetic-grasp-map} shows examples of synthetic grasp map generated by this function for three end-effectors.

\begin{algorithm}%[hbt!]
	\caption{Synthetic Experiments}\label{alg:mpc}
	\begin{algorithmic}[1]
	\Function{GenerateGraspModel}{$\langle H, W \rangle, \{p^i\}^m_{i=1}$}
	\State {\bf Inputs}: $\langle H, W \rangle$ are the height and the width of the grasp map, $m$ positions $p^i = \langle p^i_x, p^i_y \rangle$ where $0 \leq p^i_x \leq W-1$ and $0 \leq p^i_y \leq H-1$ (mimicking object positions)
	\State set $\mc{G} \gets \mathbf{1}_{H \times W}$
	\For {$i=1, \ldots, m$}
	    \State Sample a random standard deviation $\sigma_i \in [0, 1]$ and a scale $\beta_i \in [0, 1]$
	    \State Update: $\mc{G} \gets \mc{G} \times \beta_i \mc{N}(p_i, \sigma_i)$
	\EndFor
	\State return $\mc{G}$
	\EndFunction\\
	\Function{RunExperiment}{$N, \langle H, W \rangle, m, n_{ee}, h, k, l, c$}
	\State {\bf Inputs}: number of episodes $N$, number of end-effectors $\langle H, W \rangle$ are the height and the width of the synthetic grasp map, $m$ number of object centers, number of end-effectors $n_{ee}$, horizon $h$, sparsity factor $k$, void radius $l$, tool changing costs $c < 0$
		\For {$i=1, 2, \ldots, N$}
		    \State Select a random end-effector $\mc{E} \in \{1, \hdots, n_{ee}\}$
		    \State Sample $m$ random object centers $\{p^i\}^m_{i=1}$
		    \For {$j=1, 2, \ldots, n_{ee}$}
		    \State $\bar{\Gamma}_j \gets \mathit{GenerateGraspModel}(\langle H, W \rangle, \{p^i\}^m_{i=1})$ 
		    \EndFor
		    \State Construct the current plan space: $\Omega \gets \cup_{j=1}^{n_{ee}} \label{alg:plan-space} \bar{\Gamma}_j$
		    \State Set state $s \gets \langle \Omega, \mc{E}\rangle$  
		    \State Run ILP($s, h, c, l$) and STS($s, h, c, l, k$), calculate and store plan value and plan time for both solvers
		\EndFor
	\EndFunction
	\end{algorithmic} 
\label{alg:synthetic-gras-map}
\end{algorithm}
In our experiments, we report the {\em advantage} metric, defined as:
\begin{equation}
    \mbox{Adv}(\mc{T}) = \mbox{ILP}(\mc{T})-\mbox{STS}(\mc{T})
    \label{eq:adv}
\end{equation}
where $\mbox{ILP}(\mc{T})$ and $\mbox{STS}(\mc{T})$ denote the return of the best plan in each algorithm calculated for horizon $H$ using the reward function defined in Equation~\ref{eq:mpc-reward}, respectively. These values represent the long-horizon performance of each algorithm. We report $\mbox{Adv}(\mc{T})$ which measures the advantage in optimization quality of $\mbox{ILP}$ over $\mbox{STS}$, and the \textit{planning time} for each algorithm, both evaluated on our Python implementation of STS, and the commercial Gurobi~\cite{gurobi} ILP solver, using MacBook Pro 2.8 GHz Quad-Core Intel Core i7 hardware. We used a fixed void radius $l=20$ and swap cost $c=-0.2$, and report results over $n=100$ random problem instances as defined above.

Figure~\ref{fig:synthetic-grasp-map} shows our results for number of end-effectors 2 (Top two rows), and 3 (Bottom two rows). In each group, the top row shows the {\em advantage} results over STS sparsity factor $k \in \{1, 2, 3\}$  and various  horizons. The bottom row shows the  planning time for each case. In terms of quality, STS is observed to perform as well as ILP or just marginally worse. In terms of planning speed, STS is orders of magnitude faster in both cases. Yet, we observe that even in this setting, reducing $k$ significantly improves speed with a negligible effect on quality. These results motivate us to use STS in our real world application.

\begin{figure}
    \centering
    \includegraphics[width=0.9\textwidth]{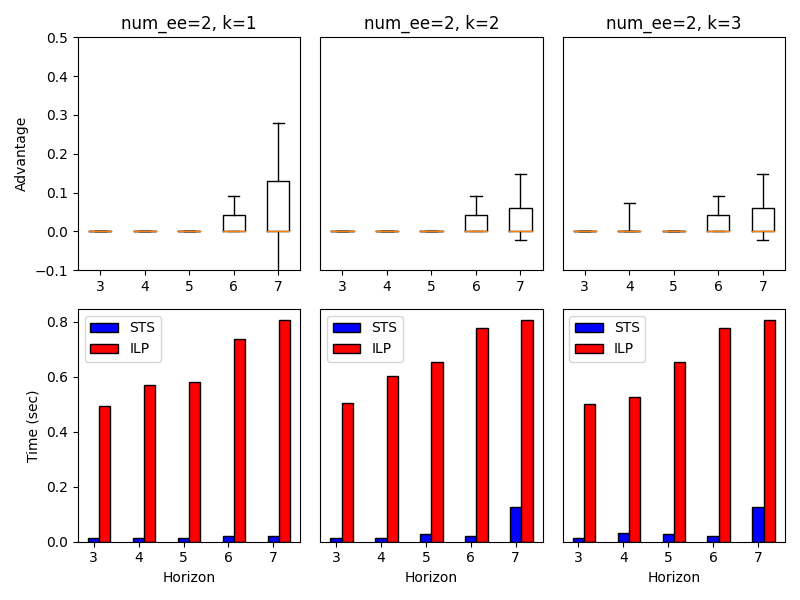}\\
    \includegraphics[width=0.9\textwidth]{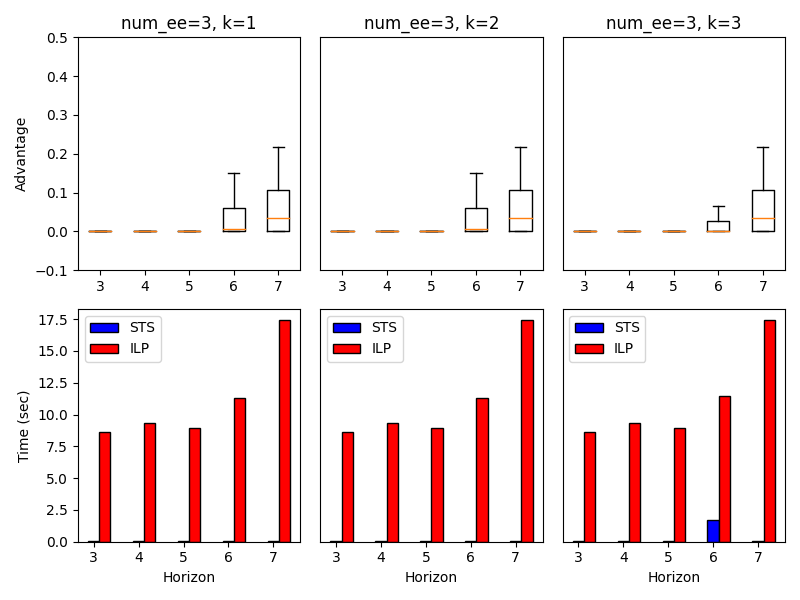}
    \caption{Performance results for synthetic experiments: (Top 2 rows) 2 end-effectors; (Bottom 2 rows) 3 end-effectors. While ILP marginally is better than STS in terms of advantage, STS yields superior speedup over ILP.}
    \label{fig:synthetic-performance}
    % \vspace{-1em}
\end{figure}

\subsection{Real World Experiments}
We conducted experiments to evaluate the performance of various grasp tool selection algorithms, and to validate the adequacy of the proposed tool changing score in capturing efficiency. First, we compare the MPC-STS with a set of heuristic baselines. Next, we compare the performance of MPC-STS
and heuristics baselines against experiments where only a single end-effector was used (no tool changing allowed). We also conduct a series of ablations on MPC-STS in terms of its void radius and max horizon (i.e., $H$). Before we present our results, we first discuss how real-world performance should be best evaluated.

% In this section we present our experimental results in a real world tool selection problem. Our goal is to evaluate the  performance improvement possible by planning the tool change based on our method, and compare with heuristic tool change strategies. As the synthetic results indicated that STS is much faster than ILP yet suffers a negligible loss in performance, we chose it in our implementation. Before we present our results, we first discuss how real-world performance should be best evaluated.
% enough for demonstrating a significant improvement, and most of our results are based on it. 
% In Sec.~\ref{ss:mpc_vs_ilp}, we also show that the ILP formulation allows to plan with the full
% grasp plan space, in contrast to MPC where we needed to sample grasp proposals during the tree search for scalability, which can further improve performance.
% Please note that most of these results benchmark the core MPC algorithm, and toward the end we also present some preliminary results using the ILP approach.
\paragraph{Metric Definitions}
\label{sec:metric-def}
Our primary goal is to  minimize the cost associated with changing tools, yet still maximize pick success. One way to measure  performance is by \emph{grasp throughput} -- the average number of successful picks in a unit time. 
% In this scheme, the cost associated with tool changing is primarily the \emph{time cost} associated with the tool change. However, pick executions over time intervals can be difficult to normalize and compare across setups, partly due to a variety of speed choices that may vary based on the robot, item size and shape, etc\AT{I'm not sure I understand this last sentence. Can you clarify? In the next sentence, in comparison, it is very clear what the difficulty is.}. In addition, 
However, grasp throughput does not correctly penalize strategies that execute many failed grasps quickly, which can be inappropriate for scenarios where items may become damaged as a result of repeated, aggressive picking.

To address this, we propose a combined score based on pick success rate (PSR), and tool consistency rate (TCR), defined as:
% these shortcomings, we propose the supplemental consideration of the following \emph{speed agnostic} metrics: pick success rate (PSR), tool consistency rate (TCR), and a combined score based on these metrics.
% \begin{equation*}
$
\small
    \mbox{PSR} = \frac{\mbox{PS}}{\mbox{PA}}, \hspace{0.2in} \mbox{TCR} = 1-\frac{\mbox{TC}}{\mbox{PA}},
    \label{eq:pick-success-rate}
$
% \end{equation*}
where $\mbox{PS}$ is the pick success count, $\mbox{PA}$ is the pick attempt count, and $\mbox{TC}$ is the tool change count (here, we do assume that there is no more than one tool change per pick attempt). Ideally, we would like both scores to be high. Also, the PSR and TCR should be balanced according to the time cost of tool change compared to the time cost of a failed grasp. We posit that the following $\beta\text{-}\mbox{TC}\text{-}\mbox{score}$ score captures these desiderata,
% of them would provide a score that is sensitive
% \begin{equation*}
%     TCR := 1-\frac{TC}{PA}
%     \label{eq:tool-consistency-rate}
% \end{equation*}
% These two readily-available metrics capture the two aspects of our desired optimization: improved pick success while being efficient about our tool changing choices. These two values are also intuitively bounded in $[0,1]$. Since it is convenient to optimize a singular metric, we propose combining these two into a single score using a harmonic mean. We prefer the harmonic mean over other means (e.g. an arithmetic mean) because it is more pessimistic than other simple means (via the HM-GM-AM inequality~\cite{bullen2013means}), particularly when performance of one of the constituent metrics is very poor in comparison to the other.
% \begin{equation*}
% \small
%     \mbox{TC}\text{-}\mbox{score} := \mbox{HarmMean}(\mbox{PSR}, \mbox{TCR}) = \frac{2 * \mbox{PSR} * \mbox{TCR}}{\mbox{PSR} + \mbox{TCR}}
%     \label{eq:tool-changer-score}
% \end{equation*}
% which is readily generalizable to control the tradeoff between these two metrics as follows:
\begin{equation}
\small
    \beta\text{-}\mbox{TC}\text{-}\mbox{score} = \frac{(1+\beta^2) * \mbox{PSR} * \mbox{TCR}}{\beta^2\mbox{PSR} + \mbox{TCR}},
    \label{eq:tool-changer-beta-score}
\end{equation}
where $\beta$ is analogous to an $F\text{-}beta$ score~\cite{baeza1999modern}. We recommend that $\beta$ be set to the \textit{opportunity cost} of a single tool change -- the approximate number successful picks that could have been completed in the time it takes to execute a tool change. 
% \AT{Why?? Where's the intuition? Also how to compute opportunity cost?}. 
For our setup, we estimated $\beta$ to be $0.33$. 

To further motivate the idea behind this metric, we present a simple numerical example that further clarifies the $\beta\text{-}\mbox{TC}\text{-}\mbox{score}$ (see Equation 3, in \textit{Real World Experiments} in the main paper): consider two tool selection algorithms A, B, being evaluated over a similar scenario (independently) with two items in the bin (i.e., each needs two successful picks to clear the bin):
\begin{equation*}
\begin{aligned}
A&: T \hspace{0.02in} F \hspace{0.02in} F \hspace{0.02in} F \hspace{0.02in} S \hspace{0.02in} T \hspace{0.02in}S \\
B&: T \hspace{0.02in} F \hspace{0.02in} F \hspace{0.02in} F \hspace{0.02in}S \hspace{0.02in} F \hspace{0.02in} F \hspace{0.02in} F \hspace{0.02in}S
\end{aligned}
\end{equation*}

In these sequences, T is a tool change event, F is pick fail, and S is pick success. Assume each pick attempt takes 1 second, and each tool change takes 3 seconds. In the above trajectories, both A and B have the same throughput (2 successes per 11 seconds). But we have a preference for A due to less failed pick attempts (A has 3 vs. 6 in B). In each case we  have:
\begin{equation*}
\begin{aligned}
A&: PSR = 2/5, TCR = 1 - 2/5 = 3/5\\
B&: PSR = 2/8, TCR = 1 - 1/8 = 7/8
\end{aligned}
\end{equation*}
For small values of $\beta$ (e.g., $\beta < 1.0$),  TC score places more importance on PSR. For extreme value $\beta = 0$, we have TCR = PSR, ignoring the cost of tool change. 

For $\beta = 0.0$, we are favoring pick success rate:
\begin{equation*}
\begin{aligned}
A&: \beta\text{-}\mbox{TC} = PSR = 0.4\\
B&: \beta\text{-}\mbox{TC} = PSR = 0.25
\end{aligned}
\end{equation*}
For larger values of $\beta$ (e.g., $\beta > 1.0$), TCR gains more importance and in the limit of $\beta = \inf$, we have TC = TCR. In the above example, for $\beta = 2$, the TC scores B higher than A.

For $\beta = 2$, we are penalizing tool changing more (by letting TCR impact the score more dominantly than PSR):
\begin{equation*}
\begin{aligned}
A&: \beta\text{-}\mbox{TC} = 5 * (2/5) * (3/5)  / (4 * (2/5) + (3/5)) = 0.545\\
B&: \beta\text{-}\mbox{TC} = 5 * (2/8) * (7/8)  / (4 * (2/8) + (7/8)) = 0.583
\end{aligned}
\end{equation*}
As we suggested in the paper, a good balance is obtained when selecting $\beta$ to be the opportunity cost. Here, the overall pick success rate is (2+2)/(5+8)~0.3, and therefore the opportunity cost is slightly less than 1.

For $\beta = 1$ we obtain:
\begin{equation*}
\begin{aligned}
A&: \beta\text{-}\mbox{TC} = 2 * (2/5) * (3/5)  / (1 * (2/5) + (3/5)) = 0.48\\
B&: \beta\text{-}\mbox{TC} = 2 * (2/8) * (7/8)  / (1 * (2/8) + (7/8)) = 0.39
\end{aligned}
\end{equation*}
Favoring A, but taking into account the cost of tool swap.

% the opportunity cost associated with a tool change to be approximately 0.33 pick attempts and, thus, selected $\beta=0.33$ for our comparisons. This singular score allows for an intuitive, decomposable measure of overall pick performance that is robust to pick speed, which may potentially vary in heterogeneous deployments, and that penalizes overly aggressive pick strategies.
% \AT{I think that this whole section could use a little more explanations behind the different mathematical choices ($\beta$ in particular). There are a lot of insights here, which may be mistakenly interpreted as ad hoc choices. Can we back them up with real world examples? (e.g., the example with execute many failed grasps quickly is very enlightening. We need more like that)}
%%%%%%%%%%%%%%%%%%%%%%%%%%%%%%%%%%%%%%%%%%%%%%%%%%%%%%%%%%%%%%%%%%%%%%%%%%%%%%%%%%%%%%%%%%%%%%%%%%%%%%%%%%%%%%%%%%%%%%%%%
%%%%%%%%%%%%%%%%%%%%%%%%%%%%%%%%%%%%%%%%%%%%%%%%%%%%%%%%%%%%%%%%%%%%%%%%%%%%%%%%%%%%%%%%%%%%%%%%%%%%%%%%%%%%%%%%%%%%%%%%%
% \vspace{-0.1in}
\paragraph{Experimental Setup}
\label{sec:experimental_setup}

We used a {\em Fanuc LRMate 200iD/7L} arm, with a tool selection hardware using two vacuum end-effectors: {\em Piab BL30-3P.4L.04AJ} (30mm) and {\em Piab BL50-2.10.05AD} (50mm). We used an assortment of mixed items (various sizes, weights, shapes, colors, etc., see Figure~\ref{fig:tc-example}(b) for an example). Each end-effector is associated with a grasp proposal model trained using previously collected production data appropriate for that end-effector. Since it is not in the scope of this paper, we only provide a brief overview of our grasp proposal model architecture. Our grasp proposal models are
inspired by the architecture proposed in~\cite{DBLP:journals/corr/abs-2006-08903} which consists of
encoder-decoder convolutional neural nets consisting of a feature pyramid network~\cite{fpn-2017} on a {\em ResNet-101} backbone and a pixelwise sigmoidal output of volume $W \times H$, where $W \times H$ are the dimensions
% of the input image (which essentially maps the input image to pixel-wise 
of the grasp success probabilities $\Gamma_{\mc{E}}$. The network is then trained end-to-end using previously collected grasp success/failure data consisting of $~5k$ grasp data per end-effector using stochastic gradient descent with momentum ($LR = 0.0003; p = 0.8$). Following the synthetic experiments conclusion, we only used the STS solver.

\begin{comment}
The feature extractor is pretrained on the MS-COCO detection task, then the full network (feature extractor with the sigmoidal output head) is trained using stochastic gradient descent with momentum ($LR = 0.0003; p = 0.8$). To compensate for the sparsity of the pixelwise success signal, we also conduct random sampling of background pixels and include their negative signals in training. Pixels are proposed from the output volume using Boltzmann sampling ($kT = 0.01$).
\end{comment}

%%%%%%%%%%%%%%%%%%%%%%%%%%%%%%%%%%%%%%%%%%%%%%%%%%%%%%%%%%%%%%%%%%%%%%%%%%%%%%%%%%%%%%%%%%%%%%%%%%%%%%%%%%%%%%%%%%%%%%%%%
%%%%%%%%%%%%%%%%%%%%%%%%%%%%%%%%%%%%%%%%%%%%%%%%%%%%%%%%%%%%%%%%%%%%%%%%%%%%%%%%%%%%%%%%%%%%%%%%%%%%%%%%%%%%%%%%%%%%%%%%%

\paragraph{Comparison with Baselines}
\label{sec:baselines}
\begin{table}[htb]
\begin{center}
\begin{small}
    \centering
    \begin{tabular}{@{}cccccc@{}}
     \toprule
     \shortstack{Algorithm \\ (w/30mm + 50mm)} & TC & PA & PS & \shortstack{TC-Score \\  ($\beta$=0.33)} & PS/hr \\ [0.5ex] 
     \midrule
     Randomized & 800 & 2191 & 744 & 0.3558 & 186 \\
     \hline
     Naive Greedy & 733 & 2093 & 1268 & 0.6099 & 317 \\ 
     \hline
     Greedy & 261 & 2702 & 1288 & 0.4999 & 295.41 \\
     \hline
     MPC-STS & 229 & 2563 & 1719 & {\bf 0.6885} & {\bf 429.75}\\
     \bottomrule
    \end{tabular}
\end{small}
    \caption{\label{tab:tool-change-alg-compare}Performance comparison over different tool selection algorithms. PS/hr is the throughput.
    % \AT{what's PS/hr and why is it different from PS?}
    }
\end{center}
\end{table}
Table \ref{tab:tool-change-alg-compare} compares our method (MPC-STS) with 3 baselines. The first 
is a {\em randomized selector}, which randomly changes tools with probability $p=0.75$ at each step, and forcing a change if not swapped after 10 steps. 
% {\em max steps}. 
The second baseline is {\em naive greedy selector}, which chooses the next grasp proposal based on one-step reward function (see Equation~\ref{eq:mpc-reward}).
% , given the current state of the bin. 
The third baseline is {\em greedy selector}, which accumulates the top $n=5$ likelihood scores for each tool, and selects the tool with the highest sum. 
% selects an end-effector that accumulates the highest sum of top $n$ available grasp proposals scores (and hence totally ignores the possibility of tool changing within the $H$ steps)\AT{This is optimized \textit{every} time step, right? Need to emphasize it then.}. 
% And finally the fourth algorithm considered in this comparison is an implementation of MPC-STS as described in Algorithm~\ref{alg:mpc-func}. For this comparison, the Randomized selector had a switch probability of $0.75$ with a maximum of 10 steps before forcing a tool change. 
% The greedy selector was configured to use the top $n=5$ grasp proposals. 
Our MPC-STS selector was configured with a void radius of $l=100$mm (roughly $60$ pixels), a maximum of $10$ initial grasp proposal samples per end-effector, sparsity factor $k=2$, and a max horizon of $H=2$ (since it yielded the best results for MPC-STS in this domain based on the ablation results in Table~\ref{tab:ablation-max-horizon}).
% \AT{2 is very short! Any explanation why? It's also strange that we didn't choose H=2 for the baselines...}. 
% For selection of the subset of grasp proposals (line~\ref{alg:mpc-subsample} in  Algorithm~\ref{alg:mpc-func}) we uniformly sampled single next best grasp proposal per end-effector. Note we used a swap reward of $c=-0.1$ in naive greedy and MPC-STS experiments. 
% The results are presented in Table~\ref{tab:tool-change-alg-compare} which shows 
Observe that MPC-STS significantly outperforms the other baselines
in terms of both TC-score and pick success rate per hour (improving over the best baseline by $50\%$).\\

\begin{comment}
The reward decay (gamma) was set to $1.0$. The swap reward is set to $-0.1$ here for both Greedy and MPC-STS based on consideration of previous empirical pick success and the time cost of swapping the tool. For example, we estimated that a tool change has a time cost $\sim 1/3$ of a pick and that, at random and without regard to the tool, we should be successful in our pick $\sim 33\%$ of the time, leading to an estimated greedy opportunity cost of $0.1$. 
\end{comment}

%%%%%%%%%%%%%%%%%%%%%%%%%%%%%%%%%%%%%%%%%%%%%%%%%%%%%%%%%%%%%%%%%%%%%%%%%%%%%%%%%%%%%%%%%%%%%%%%%%%%%%%%%%%%%%%%%%%%%%%%%
%%%%%%%%%%%%%%%%%%%%%%%%%%%%%%%%%%%%%%%%%%%%%%%%%%%%%%%%%%%%%%%%%%%%%%%%%%%%%%%%%%%%%%%%%%%%%%%%%%%%%%%%%%%%%%%%%%%%%%%%%

\paragraph{Single end-effector Comparison}
\label{sec:single-eef}

This set of comparisons is based on a separate set of shorter experimental runs with similar items; results are reported in Table~\ref{tab:single-eef-comparison}.
% \AT{do you mean items?} used in Section~\ref{sec:baselines}. 
Here, note the divergence between the TC-score and the throughput (PS/hr) in the ordering of the performance of the single $50$mm end-effector and the naive greedy baseline.
% (Table~\ref{tab:single-eef-comparison}). 
\begin{table}[htb]
\begin{center}
\begin{small}
    \centering
    \begin{tabular}{@{}cccccc@{}}
     \toprule
     Configuration & TC & PA & PS & \shortstack{TC-Score \\ ($\beta=0.33$)} & PS/hr\\ [0.5ex] 
     \midrule
     Single (30mm) & 0 & 745 & 359 & 0.508 & 287.2\\ 
     \hline
     Single (50mm) & 0 & 864 & 572 & 0.685 & 490.3\\
     \hline
     Naive Greedy & 217 & 636 & 465 & 0.751 & 348.8\\
     (30mm + 50mm) &  &  &  &  & \\
     \hline
     MPC-STS & 71 & 691 & 524 & {\bf 0.770} & {\bf 507.1}\\
     (30mm + 50mm) &  &  &  & & \\
     \bottomrule
    \end{tabular}
\end{small}
    \caption{\label{tab:single-eef-comparison}Comparison of single end-effector performance vs multiple end-effectors and tool selection.}
    % \vspace{-2em}
    \end{center}
\end{table}
While the throughput for the single $50$mm strategy is higher, the TC-score correctly reflects that this strategy is less pick efficient. Indeed, the successful pick percentage for the $50$mm strategy is ~$66\%$ while the successful pick percentage for the naive greedy strategy is ~$73\%$. The throughput in this case is inflated by executing failing picks quickly. 
% However, this strategy is undesirable in many cases, particularly if items can be damaged or flipped out of the bin during execution. Throughput, which is necessarily tied to execution time, neglects pick efficiency in this case while the proposed tool changing score correctly does not. 
As expected, MPC-STS outperforms all the baselines.

%%%%%%%%%%%%%%%%%%%%%%%%%%%%%%%%%%%%%%%%%%%%%%%%%%%%%%%%%%%%%%%%%%%%%%%%%%%%%%%%%%%%%%%%%%%%%%%%%%%%%%%%%%%%%%%%%%%%%%%%%
%%%%%%%%%%%%%%%%%%%%%%%%%%%%%%%%%%%%%%%%%%%%%%%%%%%%%%%%%%%%%%%%%%%%%%%%%%%%%%%%%%%%%%%%%%%%%%%%%%%%%%%%%%%%%%%%%%%%%%%%%

\paragraph{Parameter Study}
\label{sec:ablations}
% Void Radius
% % Discount Factor
% \begin{table}[H]
%     \centering
%     \begin{tabular}{@{}cccccc@{}}
%      \toprule
%      \shortstack{MPC Configuration \\(H = 3, M=100)} & TC & PA & PS & \shortstack{TC-Score \\ (\beta=0.33)} & PS/hr\\ [0.5ex]
%      \midrule
%      G = 0.94 & 61 & 639 & 472 & 0.752 & 472.0\\ 
%      \hline
%      G = 0.97 & 90 & 775 & 696 & 0.897 & 596.6\\
%      \hline
%      G = 1.0 & 58 & 649 & 431 & 0.682 & 417.1\\
%      \bottomrule
%     \end{tabular}
%     \caption{\label{tab:ablation-reward}Ablation over discount factor.}
% \end{table}
In these experiments, reported in Tables \ref{tab:ablation-void-radius} and \ref{tab:ablation-max-horizon}, we investigate the dependence on the void radius and  max horizon.
% each show non-trivial effects on performance. 
\begin{table}[htb]
\begin{center}
\begin{small}
    \centering
    \begin{tabular}{@{}cccccc@{}}
     \toprule
     \shortstack{MPC-STS \\ (H=3, k=2)} & TC & PA & PS & \shortstack{TC-Score \\ ($\beta=0.33$)} & PS/hr\\ [0.5ex]
     \midrule
     l = 50mm & 72 & 720 & 586 & {\bf 0.822} & {\bf 540.9} \\ 
     \hline
     l = 100mm & 58 & 649 & 431 & 0.682 & 417.1 \\
     \hline
     l = 150mm & 98 & 619 & 409 & 0.675 & 348.8 \\
     \bottomrule
    \end{tabular}
    \end{small}
    \caption{\label{tab:ablation-void-radius}Investigation of void radius $l$ (in mm).}
    % \vspace{-1em}
    \end{center}
\end{table}
% max horizon
On our item set, increasing the size of the void radius leads to a decrease in tool-changing efficiency and overall throughput at an MPC-STS with $H=3$. As the tree search progresses, the bin becomes increasingly voided. For large void radii, a large fraction of the bin will be voided, leading to unreliable reward estimates. 

Thus, as long as the void radius is large enough to cover areas disturbed by previous picks, the smaller radius the better.
% If too much of the bin is voided, estimates of rewards may become increasingly unreliable. In that case, it seems reasonable that there should be more performance at smaller void radii, insofar as the void radius is able to cover areas disturbed by previous picks.
We also see that increasing the max horizon $H$ from $1$ to $2$ leads to an increase in performance, but thereafter there is a decrease in performance metrics even though the overall tool change count remains similar. 
\begin{table}[htb]
\begin{center}
\begin{small}
    \centering
    \begin{tabular}{@{}cccccc@{}}
     \toprule
     \shortstack{MPC-STS \\ ($k=2$, $l$=100mm)} & TC & PA & PS & \shortstack{TC-Score \\ ($\beta=0.33$)} & PS/hr\\ [0.5ex]
     \midrule
     H = 1 & 64 & 712 & 522 & 0.747 & 481.8\\ 
     \hline
     H = 2 & 60 & 653 & 511 & {\bf 0.793} & {\bf 502.6}\\
     \hline
     H = 3 & 58 & 649 & 431 & 0.682 & 417.1\\
     \hline
     H = 5 & 65 & 646 & 365 & 0.586 & 353.2\\
     \bottomrule
    \end{tabular}
\end{small}
    \caption{\label{tab:ablation-max-horizon}Investigation of planning horizon.}
% \vspace{-3em}
\end{center}
\end{table}
We conjecture that this is due the crude approximation of the deterministic dynamics, which are not reliable for a long planning horizon.

\section{Conclusions and Future Directions}
\label{sec:conclusions}
In this work we introduced the Grasp Tool Selection Problem (GTSP), and presented several approximate solutions that can be deployed in real time on realistic robotic setups. Our experiments demonstrated that significant gains can be reaped by carefully planning the tool selection. For industrial bin picking, where every performance gain is directly translated to revenue, we believe that our method could be valuable.
% as an approximate MDP formulation suitable for planning based on MPC. We also presented an integer linear program formulation of the problem, and introduced a set of novel metrics for benchmarking tool selection algorithms. 

Deep learning based prediction models are becoming increasingly popular in robotics. Our work explored an optimization-based approach for maximizing the utilization of the learned models. In general, we believe that optimally choosing between several learned models could be relevant for other robotic tasks, for example, choosing between different gaits in robotic locomotion. The ideas in this work may inspire algorithms for more general problems.

\section*{Acknowledgements}
Aviv Tamar is funded by the European Union (ERC, Bayes-RL, Project Number 101041250). Views and opinions expressed are however those of the author(s) only and do not necessarily reflect those of the European Union or the European Research Council Executive Agency. Neither the European Union nor the
granting authority can be held responsible for them.
%===============================================================================

\bibliographystyle{alpha}
\bibliography{gtsp-technical-report}

\newcommand{\etalchar}[1]{$^{#1}$}
\begin{thebibliography}{BYRN{\etalchar{+}}99}

\bibitem[Ber12]{bertsekas2012dynamic}
Dimitri Bertsekas.
\newblock {\em Dynamic programming and optimal control: Volume I}, volume~1.
\newblock Athena scientific, 2012.

\bibitem[BYRN{\etalchar{+}}99]{baeza1999modern}
Ricardo Baeza-Yates, Berthier Ribeiro-Neto, et~al.
\newblock {\em Modern information retrieval}, volume 463.
\newblock ACM press New York, 1999.

\bibitem[CA13]{camachoMPC2013}
E.~F. Camacho and C.~B. Alba.
\newblock Model predictive control.
\newblock In {\em Springer Science \& Business Media}, 2013.

\bibitem[DFJ54]{danzig1954}
George~Bernard Dantzig, D.~R. Fulkerson, and Selmer~Martin Johnson.
\newblock {\em Solution of a Large-Scale Traveling-Salesman Problem}.
\newblock RAND Corporation, Santa Monica, CA, 1954.

\bibitem[EFD{\etalchar{+}}18]{ebert2018visual}
Frederik Ebert, Chelsea Finn, Sudeep Dasari, Annie Xie, Alex Lee, and Sergey
  Levine.
\newblock Visual foresight: Model-based deep reinforcement learning for
  vision-based robotic control, 2018.

\bibitem[EFLL17]{ebert2017selfsupervised}
Frederik Ebert, Chelsea Finn, Alex~X. Lee, and Sergey Levine.
\newblock Self-supervised visual planning with temporal skip connections, 2017.

\bibitem[FL16]{finn2016}
Chelsea Finn and Sergey Levine.
\newblock Deep visual foresight for planning robot motion.
\newblock {\em CoRR}, abs/1610.00696, 2016.

\bibitem[GKR20]{DBLP:journals/corr/abs-2006-08903}
Ben Goodrich, Alex Kuefler, and William~D. Richards.
\newblock Depth by poking: Learning to estimate depth from self-supervised
  grasping.
\newblock {\em CoRR}, abs/2006.08903, 2020.

\bibitem[{Gur}22]{gurobi}
{Gurobi Optimization, LLC}.
\newblock {Gurobi Optimizer Reference Manual}, 2022.

\bibitem[LDG{\etalchar{+}}17]{fpn-2017}
T.~{Lin}, P.~{Dollár}, R.~{Girshick}, K.~{He}, B.~{Hariharan}, and
  S.~{Belongie}.
\newblock Feature pyramid networks for object detection.
\newblock In {\em 2017 IEEE Conference on Computer Vision and Pattern
  Recognition (CVPR)}, pages 936--944, 2017.

\bibitem[LLS15]{lenz2015deep}
Ian Lenz, Honglak Lee, and Ashutosh Saxena.
\newblock Deep learning for detecting robotic grasps.
\newblock {\em The International Journal of Robotics Research},
  34(4-5):705--724, 2015.

\bibitem[LPKQ16]{levine2016learning}
Sergey Levine, Peter Pastor, Alex Krizhevsky, and Deirdre Quillen.
\newblock Learning hand-eye coordination for robotic grasping with deep
  learning and large-scale data collection, 2016.

\bibitem[MMS{\etalchar{+}}19]{Mahlereaau4984}
Jeffrey Mahler, Matthew Matl, Vishal Satish, Michael Danielczuk, Bill DeRose,
  Stephen McKinley, and Ken Goldberg.
\newblock Learning ambidextrous robot grasping policies.
\newblock {\em Science Robotics}, 4(26), 2019.

\bibitem[MTZ60]{mtz1960}
C.E. Miller, A.W. Tucker, and R.A. Zemlin.
\newblock Integer programming formulations and traveling salesman problems.
\newblock {\em Journal of Association for Computing Machinery}, 7:326–329,
  1960.

\bibitem[RA15]{redmon2015realtime}
Joseph Redmon and Anelia Angelova.
\newblock Real-time grasp detection using convolutional neural networks, 2015.

\bibitem[RN95]{russell1995artificial}
Stuart~Jonathan Russell and Peter Norvig.
\newblock Artificial intelligence: A modern approach.
\newblock 1995.

\bibitem[TTZ{\etalchar{+}}16]{tamar2016}
Aviv Tamar, Garrett Thomas, Tianhao Zhang, Sergey Levine, and Pieter Abbeel.
\newblock Learning from the hindsight plan - episodic {MPC} improvement.
\newblock {\em CoRR}, abs/1609.09001, 2016.

\bibitem[XELF19]{xie2019improvisation}
Annie Xie, Frederik Ebert, Sergey Levine, and Chelsea Finn.
\newblock Improvisation through physical understanding: Using novel objects as
  tools with visual foresight, 2019.

\bibitem[ZSY{\etalchar{+}}18]{zeng2018robotic}
Andy Zeng, Shuran Song, Kuan-Ting Yu, Elliott Donlon, Francois~R Hogan, Maria
  Bauza, Daolin Ma, Orion Taylor, Melody Liu, Eudald Romo, et~al.
\newblock Robotic pick-and-place of novel objects in clutter with
  multi-affordance grasping and cross-domain image matching.
\newblock In {\em 2018 IEEE international conference on robotics and automation
  (ICRA)}, pages 3750--3757. IEEE, 2018.

\end{thebibliography}

\end{document}